# New S-norm and T-norm Operators for Active Learning Method

Ali Akbar Kiaei[1]  Saeed Bagheri Shouraki[2]  Seyed Hossein Khasteh[1]  Mahmoud Khademi[3]  Ali Reza Ghatreh Samani[2]

[1] Artificial Creatures Lab, Sharif University of Technology, Tehran, Iran
[2] Department of Electrical Engineering, Sharif University of Technology, Tehran, Iran
[3] Digital Signal Processing Lab, Sharif University of Technology, Tehran, Iran

*kiaei@ce.sharif.edu*  *bagheri-s@sharif.edu*  *H_khasteh@ce.sharif.edu*  *Khademi@ce.sharif.edu*  *AliRezaSamany@ee.sharif.edu*

***Abstract:*** Active Learning Method (ALM) is a soft computing method used for modeling and control based on fuzzy logic. All operators defined for fuzzy sets must serve as either fuzzy S-norm or fuzzy T-norm. Despite being a powerful modeling method, ALM does not possess operators which serve as S-norms and T-norms which deprive it of a profound analytical expression/form. This paper introduces two new operators based on morphology which satisfy the following conditions: First, they serve as fuzzy S-norm and T-norm. Second, they satisfy Demorgans law, so they complement each other perfectly. These operators are investigated via three viewpoints: Mathematics, Geometry and fuzzy logic.

***Key-words:*** Active Learning Method; Ink Drop Spread; Hit or Miss Transform; Fuzzy connectives and aggregation operators; Fuzzy inference systems

## 1. Introduction

Active Learning Method [1] is a powerful recursive fuzzy modeling method without computational complexity. The main idea behind ALM is to break M.I.S.O. system into S.I.S.O. subsystems and aggregate the behavior of subsystems to obtain the final output. This idea resembles the brain activity which stores the behavior of data instead of the exact values of them. Each S.I.S.O. subsystem is expressed as a data plane (called IDS plane) resulted from the projection of the gathered data on each input-output plane. Two types of information can be extracted from an IDS plane. One is the behavior of output with respect to each input variable which is described by a curve called narrow path. The other one is the level of confidence for each input variable which is proportional to the reciprocal of variance of data around narrow path.

Narrow paths are estimated by applying Ink Drop Spread (IDS) on data points and Center of Gravity (COG) on data planes. IDS and COG are two main operators of ALM. Because these operators do not have fuzzy S-norm and T-norm properties, they fail to satisfy logical completeness criterion. These properties lead to the definition of basic operators which can aid to define other operators.

This paper introduces two new operators based on mathematical morphology. The operators serve as S-norm and T-norm. Moreover, they form a dual system of operators.

As you can see, Section 2 reviews the concept of fuzzy S-norm and T-norm. In section 3, the operators of ALM are restated and their drawbacks are declared. Two morphological algorithms are discussed in section 4 which are proven to satisfy Demorgans law. In section 5 proposed operators are represented as generalized versions of the two morphological algorithms discussed before. In section 6 the results are prepared in a comparison with Takagi-Sugeno. Finally, Conclusion is declared in section 7.

## 2. Fuzzy S-norm and Fuzzy T-norm

In fuzzy set theory, operator * is an S-norm, if it satisfies four conditions:
Commutativity: $x * y = y * x$
- Monotony: $if\ x \leq y \Rightarrow x * z \leq y * z$
- Associativity: $x * (y * z) = (x * y) * z$
- Neutrality of 0: $0 * x = x\ for\ x \in [0,1]$

Besides, * is a T-norm, if it satisfies:
Commutativity: $x * y = y * x$
- Monotony: $if\ x \leq y \Rightarrow x * z \leq y * z$
- Associativity: $x * (y * z) = (x * y) * z$
- Neutrality of 1: $1 * x = x\ for\ x \in [0,1]$

For example, minimum is a T-norm, because:
$$Min(x, y) = min(y, x)$$
$$If\ x<y\ \Rightarrow\ min(x, z) < min(y, z)$$
$$Min(x, min(y, z)) = min(min(x, y), z)$$
$$Min(1, x) = x\quad for\ x\ in\ [0, 1]$$

And Max is an S-norm, because:
$$Max(x, y) = Max(y, x)$$
$$\text{If } x<y \Rightarrow Max(x, z) < Max(y, z)$$
$$Max(x, Max(y, z)) = Max(Max(x, y), z)$$
$$Max(0, x) = x \quad \text{for x in } [0, 1]$$

# 3. Original operators of ALM

Flowchart of ALM is shown in Fig.3.1. Two operators are used to diffuse and fuse data in the space, IDS and COG respectively. IDS spreads the information in the problem space and COG extracts the behavior of diffused data. These operators act as a fuzzy curve fitting method [1]. They search for possible continuous paths by interpolating data points on data planes using a fuzzy method.

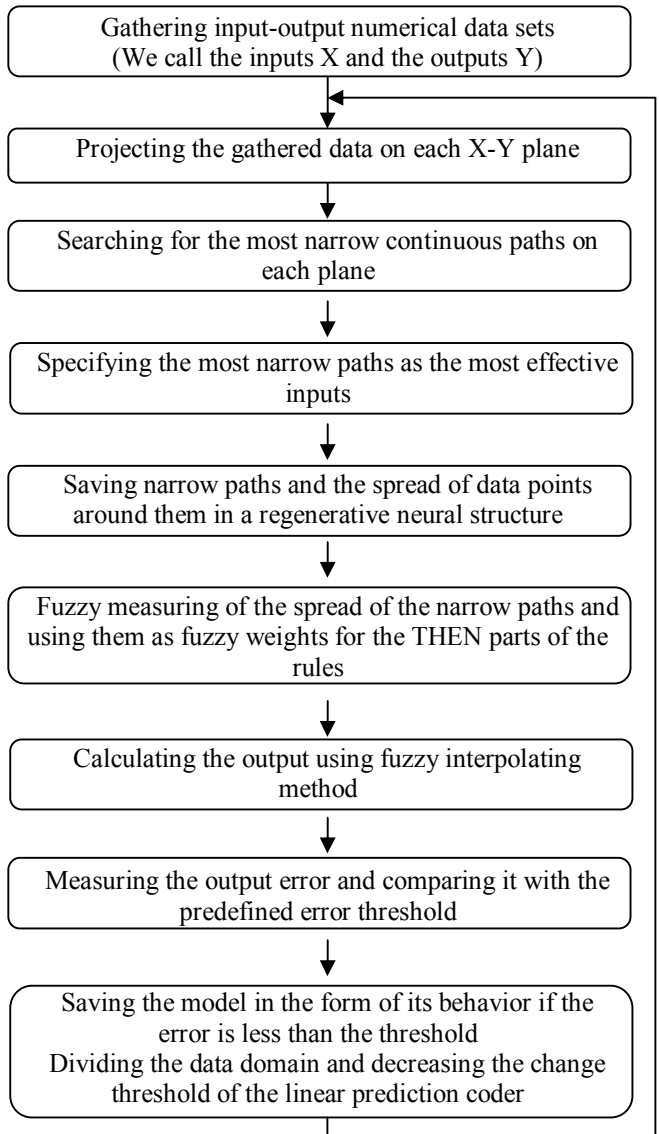

-Figure 3.1: proposed algorithm for Active Learning Method [1]

In the following both operators, IDS and COG, are described.

## 3.1. IDS

IDS considers each data point on a data plane as a light source with a cone-shaped illumination pattern. This concept is illustrated in Fig.3.2, where we have used pyramids instead of cones. The projection of process on the plane is called Ink Drop Spread. Pyramids can be considered as 2-dimensional fuzzy membership functions which are centered on each data point and show the degree of belief we have in the value of that data point and its neighbors.

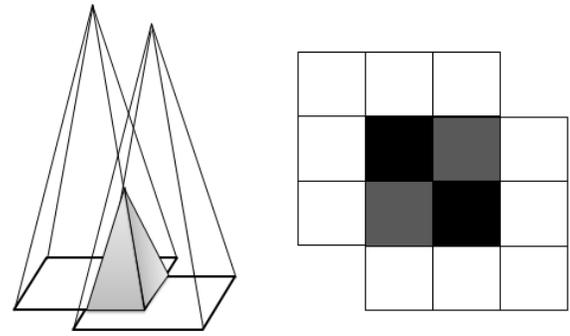

-Figure 3.2: ink drop spread and fuzzy membership functions.

Applying the algorithm on the data used by Takagi and Sugeno [23], results in Fig.3.3 for spread radiuses equal to 0.3 and 1.

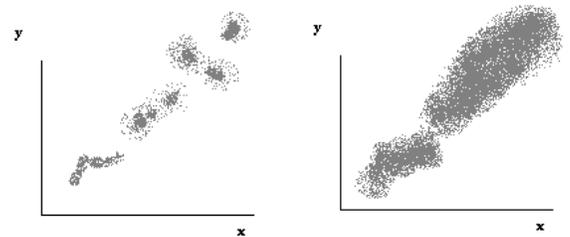

-Figure 3.3: ink drop spread results for spread radiuses equal to 0.3 and 1.

IDS satisfies fuzzy S-norm conditions. We considered each data point on a data plane is a light source which has a pyramid shape illumination pattern. Let us assume that A and B are data points in a plane. As shown in Fig.3.4, IA is the area of diffused point A and IB is the area of diffused point B using IDS.

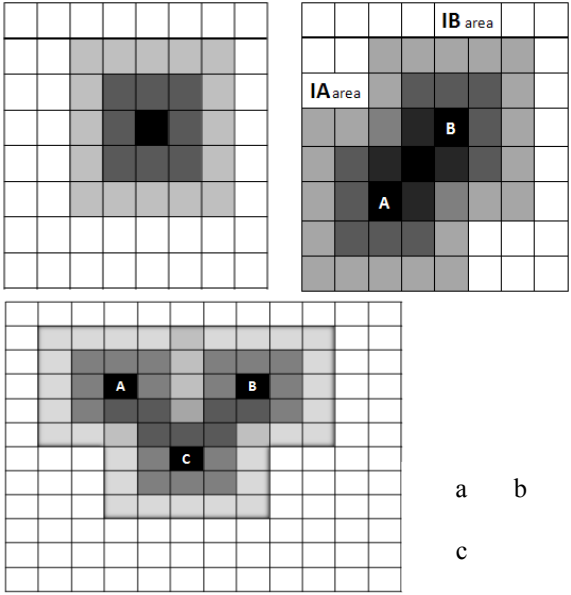
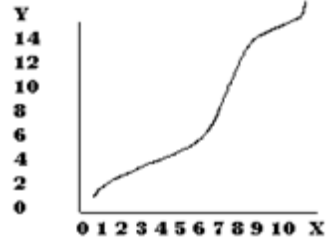

-Figure 3.4: IDS's top view on: a) one point b) two points c) three points

Here is the proof for satisfaction of S-norm conditions by IDS.

*Commutativity*: The result of applying IDS on points A and B, consecutively, is same as applying IDS on them in inverse order. Because the value of overlapped points is sum of values of all data diffused on that point, and plus operator is commutative, IDS is commutative.

*Monotony*: If the illumination of A is less than or equal to B, then all data points in IA have less or equal illumination with respect to the corresponding data points of IB. Thus for any point such as z, the illumination resulted from A is "z+ aA" which "a" is a term proportional to inverse of distance. On the other hand, the illumination of z resulted from B is "z+ aB". Since a>0, the illumination of z resulted from A is less than or equal to the illumination of z resulted from B.

*Associativity:* Assume that light sources A, B and C are going to affect to the point "x" by IDS. For each source, IDS increases the illumination with respect to distance regardless of other sources. For instance, imagine point "x" in Fig.3.4.c. The order of applying IDS on this point does not affect the distance. The value of the point "x" is sum of effects of three sources on this point. Since plus operator is associative, IDS becomes associative as well.

*Neutrality of 0:* Assume a pyramid with height 0, summation of this pyramid with others does not affect them. Hence, zero is the neutral element of IDS.

## 3.2. Center of Gravity

The Center of Gravity method, tries to select a representative point in each column of data. This point can be obtained by calculating the weighted average of all diffused data points in each column. Fig.3.5 illustrates the extracted path using COG on the plane shown previously in Fig.3.3.

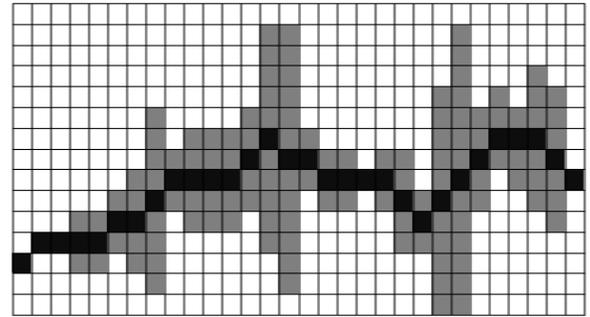

-Figure 3.5: Extracted narrow path by Center of Gravity

Another example of COG can be seen in Fig.3.6. Each column may have a different number of data points and the representative point in each one is colored in black.

-Figure 3.6: Center of Gravity in top view

COG does not satisfy Associativity. Assume that A, B, and C are three points which are all in the same column as shown in Fig.3.7. The first trial applies COG on C and the result of COG on A and B, Fig.3.7.a. On the other hand, the second trial applies COG on A and the result of COG on B and C.Fig.3.7.b.

As shown in bellow, the final results are not necessarily the same. Thus the order of actions affects the final result and Center of Gravity is not associative. Consequently COG cannot serve as a T-norm.

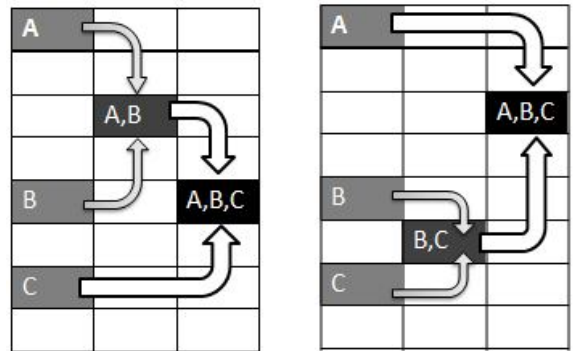

-Figure 3.7: illustrating Non-Associativity of COG

# 4. Two basic Morphological Algorithms

This section expresses two new algorithms that act like operators of ALM, and it is shown that they are dual of each other.

## 4.1. Thinning

The Thinning [8] of one set, like A, by one structure element such as B, which is shown as "$A \otimes B$", is defined as follows:
$$A \otimes B = A - (A \circledast B)$$
$$= A \cap (A \circledast B)^c \quad 4.1.1$$

The Hit or Miss Transform (HOM) [8] in the formula above is defined as:
$$A \circledast B = (A \ominus B1) \cap (A^c \ominus B2)$$

and Erosion [8] in the formula above is defined as:
$$A \ominus B = \{ c \in Z^2 | c + b \in A, \forall b \in B\}$$

where B = (B1, B2). [8]

An applicable phrase for Thinning is based on chain of structuring elements:
$$\{B\} = \{B^1, B^2, \ldots, B^n\} \quad 4.1.2$$

where $B^i$ is a rotated version of $B^{i-1}$. Considering these elements, we now redefine Thinning in this manner:
$$A \otimes \{B\} = ((\ldots(A \otimes B^1) \otimes B^2) \ldots) \otimes B^n) \quad 4.1.3$$

First A is thinned by $B^1$, and then the result is thinned by $B^2$, and so forth. At last A is thinned by $B^n$. The whole action is repeated until no change occurs. [8]

| 0 | 0 | 0 |
|---|---|---|
| * | 1 | * |
| 1 | 1 | 1 |

$B^1$

| * | 0 | 0 |
|---|---|---|
| 1 | 1 | 0 |
| 1 | 1 | * |

$B^2$

| 1 | * | 0 |
|---|---|---|
| 1 | 1 | 0 |
| 1 | * | 0 |

$B^3$

| 1 | 1 | * |
|---|---|---|
| 1 | 1 | 0 |
| * | 0 | 0 |

$B^4$

| 1 | 1 | 1 |
|---|---|---|
| * | 1 | * |
| 0 | 0 | 0 |

$B^5$

| * | 1 | 1 |
|---|---|---|
| 0 | 1 | 1 |
| 0 | 0 | * |

$B^6$

| 0 | * | 1 |
|---|---|---|
| 0 | 1 | 1 |
| 0 | * | 1 |

$B^7$

| 0 | 0 | * |
|---|---|---|
| 0 | 1 | 1 |
| * | 1 | 1 |

$B^8$

-Figure 4.1: Thinning structure elements

## 4.2. Thickening

The Thickening [8] of one set, like A, by one structure element such as B, which is shown as "$A \odot B$", is defined as follows:
$$A \odot B = A \cup (A \circledast B) \quad 4.2.1$$

Thickening can be shown as sequential operations:
$$\{B\} = \{B^1, B^2, \ldots, B^n\} \quad 4.2.2$$
$$A \otimes \{B\} = ((\ldots(A \otimes B^1) \otimes B^2) \ldots) \otimes B^n) \quad 4.2.3$$

The structuring elements [8] used for Thickening illustrated in Fig.4.2 have the same form as those shown in Fig.4.1 used for Thinning, but with all ones and zeros interchanged.

| 1 | 1 | 1 |
|---|---|---|
| * | 0 | * |
| 0 | 0 | 0 |

$B^1$

| * | 1 | 1 |
|---|---|---|
| 0 | 0 | 1 |
| 0 | 0 | * |

$B^2$

| 0 | * | 1 |
|---|---|---|
| 0 | 0 | 1 |
| 0 | * | 1 |

$B^3$

| 0 | 0 | * |
|---|---|---|
| 0 | 0 | 1 |
| * | 1 | 1 |

$B^4$

| 0 | 0 | 0 |
|---|---|---|
| * | 0 | * |
| 1 | 1 | 1 |

$B^5$

| * | 0 | 0 |
|---|---|---|
| 1 | 0 | 0 |
| 1 | 1 | * |

$B^6$

| 1 | * | 0 |
|---|---|---|
| 1 | 0 | 0 |
| 1 | * | 0 |

$B^7$

| 1 | 1 | * |
|---|---|---|
| 1 | 0 | 0 |
| * | 0 | 0 |

$B^8$

-Figure 4.2: Thickening structure elements

## 4.3. Thinning, Thickening and Demorgans law

In this section, it is shown that Thickening is the morphological dual of the Thinning.

The dual of Thinning is:
$$\left(\underbrace{A^c \otimes B^c}_{\mathbb{1}}\right)^c = ? \quad 4.3.1$$

To calculate the other side of this equation, we compute $\mathbb{1}$ with respect to eq. 4.1.1. The $\mathbb{1}$ equals to:
$$\mathbb{1} = A^c \otimes B^c = A^c \cap \left(\underbrace{A^c \circledast B^c}_{\mathbb{2}}\right)^c \quad 4.3.2$$

Based on the HOM definition [8], $\mathbb{2}$ is rewritten as:
$$\mathbb{2} = A^c \circledast B^c = (A^c \ominus B^c) \cap (A^{cc} \ominus B^{cc})$$
$$= (A^c \ominus B^c) \cap (A \ominus B)$$
$$= (A \ominus B) \cap (A^c \ominus B^c) = A \circledast B$$
$$\Rightarrow A^c \circledast B^c = A \circledast B \quad 4.3.3$$

So HOM of the elements are identical to their complement with the same action. Based on Thinning definition, we have:
$$\Rightarrow \mathbb{1} = A^c \otimes B^c = A^c \cap (A^c \circledast B^c)^c$$

$$\stackrel{4.3.3}{\iff} A^c \cap \underbrace{(A \circledast B)^c}_{③} \qquad 4.3.4$$

In the last equation, we used eq. 4.3.3, to omit the extra relations. By expanding ③, inserting it into ① and calculating the complement of ①, the dual of Thinning will be obtained.

$$③ = (A \circledast B)^c = [(A \ominus B) \cap (A^c \ominus B^c)]^c$$
$$= \underbrace{(A \ominus B)^c}_{④} \cup (A^c \ominus B^c)^c \qquad 4.3.5$$

Thus, finding complement of HOM is simplified to the finding of complement of Erosion, ④.

If we define Dilation [8] as
$$A \oplus B = \{x \mid (\hat{b})_x \cap A \neq \emptyset\}$$

Then the complement of Erosion is rewritten respect to Dilation:
$$④ = (A \ominus B)^c = \{z \mid (B_z) \subseteq A\}^c$$
$$= \{z \mid B_z \cap A^c = \emptyset\}^c$$
$$= \{z \mid B_z \cap A^c \neq \emptyset\} = A^c \oplus \hat{B} \quad 4.3.6$$

Inserting ④, ③, ② in their places, ① expands in the following way:
$$\Rightarrow ① = A^c \cap [(A^c \oplus \hat{B}) \cup (A \oplus \hat{B}^c)]$$
$$= [A^c \cap (A^c \oplus \hat{B})] \cup [A^c \cap (A \oplus \hat{B}^c)]$$

And the dual of Thinning is:
$$\Rightarrow (A^c \otimes B^c)^c = ①^c =$$
$$= \{[A^c \cap (A^c \oplus \hat{B})] \cup [A^c \cap (A \oplus \hat{B}^c)]\}^c$$
$$= [A^c \cap (A^c \oplus \hat{B})]^c \cap [A^c \cap (A \oplus \hat{B}^c)]^c$$
$$= [A \cup (A^c \oplus \hat{B})^c] \cap [A \cup (A \oplus \hat{B}^c)^c]$$
$$= [A \cup (A \ominus B)] \cap [A \cup (A^c \ominus B^c)]$$
$$= A \cup \{(A \ominus B) \cap (A^c \ominus B^c)\}$$
$$= A \odot B$$

$$\Rightarrow (A^c \otimes B^c)^c = A \odot B \qquad 4.3.7$$

Thus, we proved that Thickening is the dual of Thinning. By proving the contrary relation below, it can be inferred that Thinning and Thickening are duals of each other:
$$(A^c \odot B^c)^c = (A \otimes B)$$

By using the previous relations, we have:
$$A^c \odot B^c = A^c \cup (A^c \circledast B^c)$$
$$= A^c \cup (A \circledast B)$$
$$(A^c \odot B^c)^c =$$
$$= A \cap [(A \ominus B)^c \cup (A^c \ominus B^c)^c]$$
$$= A - [(A \ominus B)^c \cup (A^c \ominus B^c)^c]^c$$
$$= A - [(A \ominus B) \cap (A^c \ominus B^c)]$$
$$= A - (A \circledast B) = A \otimes B \qquad 4.3.8$$

Therefore, Thinning and Thickening are duals of each other.

# 5. Proposed new morphological operators

We proved in section 4 that and Thickening can serve as duals for each other, but they are not commutative. Thus, they cannot be fuzzy S-norm and T-norm. This section proposes two new operators which are in fact generalized forms of Thinning and Thickening algorithms. We called these new operators Extended Thinning and Extended Thickening respectively.

## 5.1. Extended Thinning

Let us assume that A and B are square Matrices. The Extended Thinning operator is defined as below:
$$A \times B$$
$$= \begin{cases} [0]_{size(A)} & if\ size(A) = size(B) \\ Save[\ Max(L(A), L(B)) \otimes min(L'(A), L'(B)), \\ \qquad Save(L'(A), L'(B))\ ] & otherwise \end{cases}$$
(5.1.1)

If one matrix, like C, is not square, we add sufficient number of '*'(don't care elements) to the end of matrix to change it into smallest square matrix containing the original *C*. *Max* and *Min* operators which were showed to be S-norm and T-norm in section 2, are defined on the size of matrix. For example, using this definition any 4*4 matrix is bigger than a 2*2 matrix. *Save (A, B)* is the matrix that saves information of A and B, with respect to their sizes. Implementation of this operator must consider the elements of matrix as strings, not as numbers. For example, when matrices A and B are as shown below:

$$A = \begin{matrix} 0 & 0 & 1 \\ 1 & 0 & 0 \\ 0 & 0 & 0 \end{matrix}, \quad B = \begin{matrix} 1 & 1 & 1 \\ 1 & 1 & 1 \\ 0 & 0 & 0 \end{matrix} \Rightarrow$$

Then the *Save* matrix is:
$$\begin{cases} Save(A, B) = \begin{matrix} 01 & 01 & 11 \\ 11 & 01 & 01 \\ 00 & 00 & 00 \end{matrix} \\ \\ Save(B, A) = \begin{matrix} 10 & 10 & 11 \\ 11 & 10 & 10 \\ 00 & 00 & 00 \end{matrix} \end{cases}$$

Or when matrices A and B are:
$$A = \begin{matrix} 0 & 1 \\ 1 & 0 \end{matrix}, \quad B = \begin{matrix} 1 & 0 & 0 \\ 0 & 0 & 0 \\ 0 & 1 & 0 \end{matrix}$$

the Save matrix has the form below:
$$Save(A, B) = Save(B, A) = \begin{matrix} 10 & 01 & 0* \\ 01 & 00 & 0* \\ 0* & 1* & 0* \end{matrix}$$

where '*' denotes a don't care element.

L(A) is a matrix that extracts left character of each element in matrix A. For instance, when matrix A is in the form below:

$$A = \begin{matrix} 01 & 01 & 11 \\ 11 & 01 & 01 \\ 00 & 00 & *0 \end{matrix}$$

yields L(A) as below:

$$L(A) = \begin{matrix} 0 & 0 & 1 \\ 1 & 0 & 0 \\ 0 & 0 & * \end{matrix}$$

In Extended Thinning, L(A) extracts a numerical matrix which is used in thinning process.

To define L′, two matrices R and T must be defined. R(A) is the right character of each element in matrix A, and T(A) contains all of the elements in A except L(A) and R(A). L′(A) is defined as:

$$L'(A) = Save(T(A), R(A)) \qquad 5.1.2$$

As an example, having A as:

$$A = \begin{matrix} 010 & 010 & 11* \\ 110 & 010 & 010 \\ 001 & 001 & 00* \end{matrix}$$

yields:

$$L(A) = \begin{matrix} 0 & 0 & 1 \\ 1 & 0 & 0 \\ 0 & 0 & 0 \end{matrix} \ ; \quad R(A) = \begin{matrix} 0 & 0 & * \\ 0 & 0 & 0 \\ 1 & 1 & * \end{matrix}$$

$$\Rightarrow T(A) = \begin{matrix} 1 & 1 & 1 \\ 1 & 1 & 1 \\ 0 & 0 & 0 \end{matrix}$$

And consequently yields:

$$L'(A) = \begin{matrix} 10 & 10 & 1* \\ 10 & 10 & 10 \\ 01 & 01 & 0* \end{matrix}$$

As another example let us consider a single element matrix such as A:

$$A = \begin{matrix} 0 & 0 & 1 \\ 1 & 0 & 0 \\ 0 & 0 & 0 \end{matrix}$$

Consequently we have:

$$L(A) = \begin{matrix} 0 & 0 & 1 \\ 1 & 0 & 0 \\ 0 & 0 & 0 \end{matrix}; \quad R(A) = \begin{matrix} 0 & 0 & 1 \\ 1 & 0 & 0 \\ 0 & 0 & 0 \end{matrix}; \quad T(A) = \emptyset$$

$$L'(A) = \begin{matrix} 0 & 0 & 1 \\ 1 & 0 & 0 \\ 0 & 0 & 0 \end{matrix}$$

The expanded application of matrix $L'$ in Extended Thinning is as described in 5.1.3.

if $B = Save(B^0, B^1, B^2, ..., B^n)$ then
$A \otimes L'(B) = ((...(A \otimes B^1) \otimes B^2) \otimes ...) \otimes B^n)$
5.1.3

## 5.2. Extended Thinning as an S-norm

In this section we prove that Extended Thinning is an S-norm. It is needed to consider properties of S-norm in two diverse cases. Assume A and B are two square matrices. In case I, size(A) ≠ size(B), and in case II, size(A) = size(B). Third square matrix, called C, is needed which can be defined in five situations of size. (See Fig.5.1)

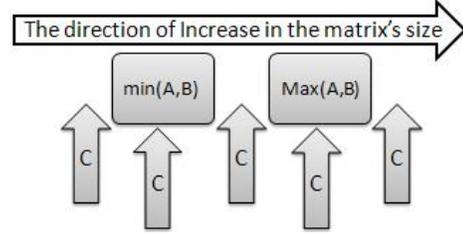

-Figure 5.1: possible positions for size of C

Case I) $size(A) \neq size(B)$

Claiming Extended Thinning to be an S-norm requires to prove that four conditions of an S-norm are satisfied: Commutativity, Monotony, Associativity and Neutrality of zero:

1. *Commutativity*

$$B \times A \overset{?}{\Leftrightarrow} A \times B$$

$A \times B = $ Save{ Max(L(A), L(B)) ⊗ min(L′(A), L′(B))
, Save(L′(A), L′(B)) }    *5.2.1*
$B \times A = $ Save{ Max(L(B), L(A)) ⊗ min(L′(B), L′(A))
, Save(L′(B), L′(A)) }    *5.2.2*

Counting Min and Max as T-norm and S-norm, we have the following relations according to section 2:
$min(A, B) = min(B, A)$    5.2.3
$Max(A, B) = Max(B, A)$    5.2.4
By rewriting eq. 5.2.2 based on eq. 5.2.3 and eq. 5.2.4
$B \times A = $ Save{ Max(L(B), L(A)) ⊗ min(L′(B), L′(A))
, Save(L′(B), L′(A)) }
$ = $ Save{ Max(L(A), L(B)) ⊗ min(L′(A), L′(B))
, Save(L′(A), L′(B)) } = $A \times B$
$\Rightarrow B \times A = A \times B$    5.2.5
Thus Commutativity is proven.

2. *Monotony*

$$A \leq B \overset{?}{\Rightarrow} A \times C \leq B \times C$$

Let us assume that $A_{a*a}$, $B_{b*b}$ and $C_{c*c}$ are three square matrixes. According to Fig. 5.1, we inspect Monotony in five separate situations:

-First situation, (c < a < b):
$$\begin{cases} A \times C = [T_1]_{size(A)} = T_{1_{a*a}} \\ B \times C = [T_2]_{size(B)} = T_{2_{b*b}} \end{cases}$$
$$\overset{a<b}{\Longrightarrow} T_1 < T_2 \quad \Rightarrow \quad A \times C < B \times C$$

-Second situation, (c = a < b):
$$\begin{cases} A \times C = [T_1]_{size(A)} = T_{1_{a*a}} \\ B \times C = [T_2]_{size(B)} = T_{2_{b*b}} \end{cases}$$
$$\overset{a<b}{\Longrightarrow} T_1 < T_2 \quad \Rightarrow \quad A \times C < B \times C$$

-Third situation, (a < c < b):
$$\begin{cases} A \times C = [T_1]_{size(c)} = T_{1_{c*c}} \\ B \times C = [T_2]_{size(B)} = T_{2_{b*b}} \end{cases}$$
$$\xRightarrow{c<b} T_1 < T_2 \Rightarrow A \times C < B \times C$$

-Fourth situation, (a < b = c):
$$\begin{cases} A \times C = [T_1]_{size(c)} = T_{1_{c*c}} \\ B \times C = [T_2]_{size(B)} = T_{2_{b*b}} \end{cases}$$
$$\xRightarrow{c=b} size(T_1) = size(T_2)$$
$$\Rightarrow A \times C = B \times C$$

-Fifth situation, (a < b < c):
$$\begin{cases} A \times C = [T_1]_{size(c)} = T_{1_{c*c}} \\ B \times C = [T_2]_{size(c)} = T_{2_{c*c}} \end{cases}$$
$$\xRightarrow{c=c} size(T_1) = size(T_2)$$
$$\Rightarrow A \times C = B \times C$$

So, in all five situations, we proved that:
if $A \leq B \Rightarrow A \times C \leq B \times C$    5.2.6
Hence, the second property of S-norm is satisfied.

3. *Associativity*
$$A \times (B \times C) \overset{?}{\Leftrightarrow} (A \times B) \times C$$
$B \times C$ is a string matrix with each element including three characters. These characters are the elements of $B \otimes C, B$ and $C$. Since the elements of B and C are single characters, $L(B) = L'(B) = B$ and $L(C) = L'(C) = C$, which consequently yield $L(B \times C) = B \otimes C$, and $L'(B \times C) = BC$.

Similar to Monotony condition, by assuming $A_{a*a}$, $B_{b*b}$ and $C_{c*c}$ as three square matrixes and considering four separate situations, we have:

-First situation, ( $a > b, c$ ):
The elements of matrix $A \times (B \times C)$, contain four characters of matrixes
$A \otimes L'(B \otimes C, B, C), A, B$ and $C$    5.2.7
Based on eq. 5.1.3, we have
$A \otimes L'(B \otimes C, B, C) = ((A \otimes B) \otimes C)$    5.2.8
Moreover, elements of $(A \times B) \times C$ contain four characters of matrixes
$( L((A \otimes B), A, B) \otimes L'(C) ), A, B$ and $C$    5.2.9
The first matrix in eq. 5.2.9 is equal to $((A \otimes B) \otimes C)$, which is the same as eq. 5.2.8.

-Second situation, ( $a < b, c$ ):
$B \times C$ is a combination of three matrixes, $B \otimes C, B$ and $C$. Since $size(A)$ is less than $size(B \otimes C)$ then :

$$L(A \times (B \times C)) = L((B \otimes C; B; C) \times A)$$
$$= [ L(B \otimes C; B; C) \otimes L'(A) ]$$
$$= (B \otimes C) \otimes A$$
Moreover, based on eq. 5.2.5 we have:
$$(A \times B) \times C = (B \times A) \times C$$
But we should also note that:
$$L((B \times A) \times C) = (B \otimes A; B; A) \times C$$
$$= [ L(B \otimes A; B; A) \otimes L'(C) ]$$
$$= (B \otimes A) \otimes C$$
Consequently two equations above are equal.

-Third situation, ( $c < a < b$ ):
The size of A is less than size $(B \otimes C)$ which yields:
$$L(A \times (B \times C)) = (B \otimes C) \otimes A$$
Moreover,
$$L((A \times B) \times C) = (B \otimes A) \otimes C$$
Hence these two equations are equal.

-Forth situation, ( $b < a < c$ ):
Because the size of matrix C is greater than B, it can be inferred that $L(B \times C) = C \otimes B$, and:
$$L(A \times (B \times C)) = (C \otimes B) \otimes A$$
In addition the following equation holds:
$$L((A \times B) \times C) = (C \otimes A) \otimes B.$$
Therefore we have:
$A \times (B \times C) = (A \times B) \times C$    5.2.10
We can finally conclude that S-norm Associativity is satisfied for all feasible situations.

4. *Neutrality of zero*
To satisfy this property it is needed to introduce *zero*.
If the center component of structure element matrix is "1", then it eliminates some part of other matrixes in Extended Thinning. Therefore, it is needed that the structure element matrix does not affect other matrixes in Extended Thinning. This matrix must have two properties:
- Its center element must be zero.
- It must be smaller than any matrix in order to become structure element.

Considering these properties, the smallest matrix with a "0" element in the center is $[0]_{1*1}$, which can serve as neutral zero for S-norm.

Based on eq. 5.2.5, 5.2.6, 5.2.10 and the paragraph above, all four conditions for being an S-norm are proven to be satisfied.

Case II) $\boldsymbol{size(A) = size(B)}$ :
Let us assume that $A_{a*a}$, $B_{b*b}$ and $C_{c*c}$ are three square matrixes. Four conditions are proven under different situations:

1.*Commutativity*
$$A \times B = [0]_{a*a} = B \times A$$

2.*Monotony*

$$if\ a = b \Rightarrow A \times C \leq B \times C$$
Assuming $Max(a,c) = d$:
$$\begin{cases} A \times C = [T_1]_{d*d} = T_{1_{d*d}} \\ B \times C = [T_2]_{d*d} = T_{2_{d*d}} \end{cases}$$
$$\xRightarrow{a=b} T_1 = T_2$$
$$\Rightarrow A \times C = B \times C$$

3. *Associativity*

Since the size of matrix A is equal to B, we analyze three situations for different sizes of matrix C:

-First situation, $(a = b > c)$ :
In this situation, $(B \times C)$ contains three matrixes $B \otimes C, B,$ and $C$ with size $b*b$. Therefore
$$A \times (B \times C) = L(A) \otimes L'(B \times C) = [0]_{a*a}$$
This is because the size of two matrices $A_{a*a}$ and $B_{b*b}$ are the same (a=b).

-Second situation, $(a = b = c)$:
Because all the three matrices have the same size, the Extended Thinning for each of them with respect to another is $[0]_{a*a}$, and consequently the final result of Extended Thinning for three matrixes is $[0]_{a*a}$.

-Third situation, $(a = b < c)$:
The size of matrix B is smaller than C, and $(B \times C)$ consists of $C \otimes B, C$ and $B$, with a size of c*c.
$$L(A \times (B \times C)) = L(B \times C) \otimes L'(A)$$
$$= (C \otimes B) \otimes A$$
Moreover, $A \times B$ contains $[0]_{a*a}, A,$ and $B$ with a size of a*a.
$$L((A \times B) \times C) = L(C) \otimes L'(A \times B)$$
$$= C \otimes L'([0], A, B) = (C \otimes A) \otimes B$$
These matrixes are equal. So, Associativity is satisfied.

4. *Neutrality of zero*

Similar to Case I, it can be proven that the neutral matrix is $[0]_{1*1}$.

Therefore, all four conditions are satisfied in all feasible situations and Extended Thinning can consequently serve as an S-norm.

## 5.3. Extended Thickening

Let us assume that A and B are square matrixes. The Extended Thickening operator is defined as:

$$A \circ B$$
$$= \begin{cases} [1]_{size(A)} & if\ size(A) = size(B) \\ Save[\ Max(L(A), L(B)) \odot min(L'(A), L'(B)), \\ \quad Save(L'(A), L'(B))\ ] & otherwise \end{cases}$$

5.3.1
All of the definitions of Max, Min, Save, L, L' and R are exactly the same as before. Hence, the proof of Commutativity, Monotony and Associativity would be the same as before. It is enough to introduce *Neutrality of one*, the fourth property of T-norm, to prove that this new operator can serve as a T-norm.

If the center element of structure element matrix is "0", then the Extended Thickening adds structure element data to the main matrix. Therefore, it is needed that the structure element matrix does not affect other matrices in Extended Thickening. This matrix must have two properties:
- Its center element must be one.
-It must be smaller than any matrix in order to become structure element.

Considering these properties, the smallest matrix with "1" in the center is $[1]_{1*1}$, which is the neutral element for a T-norm.

Therefore, we proposed two new operators that can serve as fuzzy S-norm and T-norm.

Next section proves the duality of these operators.

## 5.4. Extended Thinning, Extended Thickening and Demorgans law

Let us assume that A and B are two square matrixes. Without affecting the generality of our proof, we assume that size of A is greater than B:
$$if\ A > B \Rightarrow Size(A) > size(B)$$
$$\Rightarrow size(A^c) > size(B^c)$$
$$\Rightarrow A^c > B^c \quad 5.4.1$$

$$\xRightarrow{5.4.1} \begin{cases} size(\ Max(A,B)\ ) = Size(\ Max(A^c, B^c)\ ) \\ size(\ min(A,B)\ ) = Size(\ min(A^c, B^c)\ ) \end{cases}$$
$$5.4.2$$

Considering equations 5.4.1 and 5.4.2 simplifies proving the duality of operators $A \times B$ and $A \circ B$ to proving the duality of $A \otimes B$ and $A \odot B$. This proof is previously explained in section 4.3 completely.

## 6. Results

The first results are obtained on a simple data plane applying both original operators of ALM.

IDS and COG are executed in 5.1. Drawbacks of COG are then shown from viewpoint of geometry.

Extended Thickening and Extended Thinning are executed in 5.2 and it is shown that they do not have disadvantages of COG.

In addition, a more complicated example is stated in 5.3. which demonstrates the strength of Extended Thinning.

## 6.1. Applying IDS and COG

Assume that the data points in the plane X-Y have a circular structure. They can also have any other structure which does not satisfy the structure of functions, ($\exists x_0: f(x_0) = y_1, f(x_0) = y_2$ and $y_1 \neq y_2$).

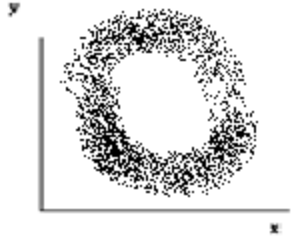

-Figure 6.1: sample data points with a structure similar to a circle.

By applying IDS and COG on this data plane, the structure of data points are ruined. It is because in each column of data plane, Center of Gravity tries to select the average as a delegate of the column.

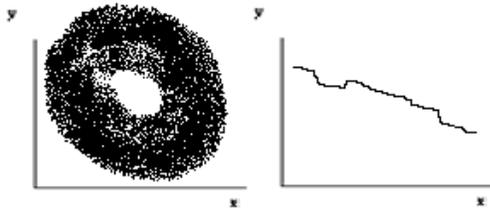

-Figure 6.2: ink drop projection and extracted narrow path by Center of Gravity

From the viewpoint of the geometry, Center of Gravity reduces a 2-dimensional space to some 1-dimensional spaces (which we previously called columns) which are located side by side. It is because COG selects only one point, called delegate point for each column in the data plane. Calculating delegates is independent of data in the neighbor columns. Since this space reduction causes destroying some pieces of information, the structure of original data points is not necessarily preserved.

## 6.2. Applying Extended Thickening and Extended Thinning

Thinning is capable of preserving the data structure. In the next stage of original ALM, data domain is divided to find narrow paths. Previously, ALM had to break narrow paths based on variance due to lack of information. On the contrary, the modified ALM divides these narrow paths by the width of thickened data in any vertical lines. Thus ALM arrives to an admissible answer faster. Using these new operators, the fuzzy rules made by ALM are more accurate and less in numbers with respect to original ALM.

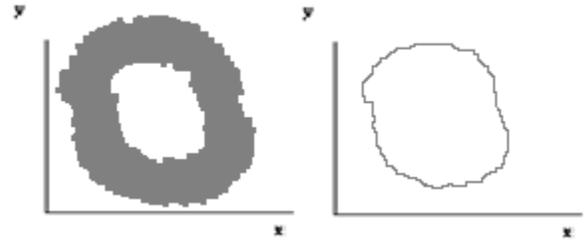

-Figure 6.3: Extended Thickening projection in the left and narrow path extracted by Extended Thinning in the right.

## 6.3. Comparing the couple operators

By giving a complicated sample, the drawback of COG will be highlighted. In this sample, data points have a structure like four chained circles and a separated half circle.

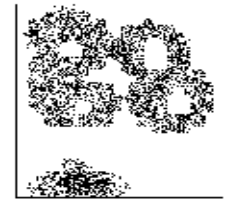

-Figure 6.4: original data points (structure consisting of four attached circles and a half circle)

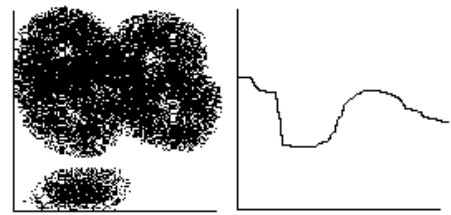

-Figure 6.5: Ink Drop Spread (IDS) and Center of Gravity applied on original data points shown in Fig. 6.4

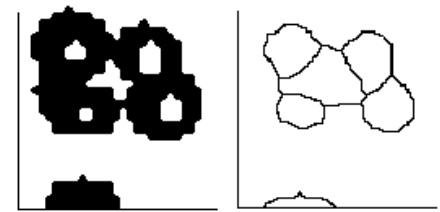

-Figure 6.6: Extended Thickening and Extended Thinning applied on original data points shown in Fig. 6.4. Notice that how the structure of input data points is protected.

In original ALM, Center of Gravity destroyed the structure of data due to its dimension reduction property. With this diminished data, original ALM had to break data points to inaccurate data sets. Now, the modified ALM divides data points based on the width of data in vertical lines. If this width in any column is bigger than a designated threshold, then two or more delegates will be chosen in this column. Consequently these delegates cannot form a function. The designated threshold is the radius of structure element.

Let us assume that the projection of data points on plane X1-Y is as shown in Figure 6.7.a. The result of applying IDS on data is shown in Fig 6.7.b.

Fig 6.8 illustrates the result of executing Center of Gravity. On the same data points, Extended Thickening is applied instead of IDS and the results are shown in Fig. 6.9. Finally the narrow path extracted by Extended Thinning is shown in Fig 6.10. It has two new excess lines compared to the result obtained by COG.

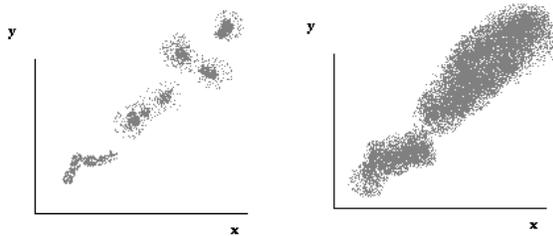

-Figure 6.7: main data points and Ink Drop Spread results (IDS)

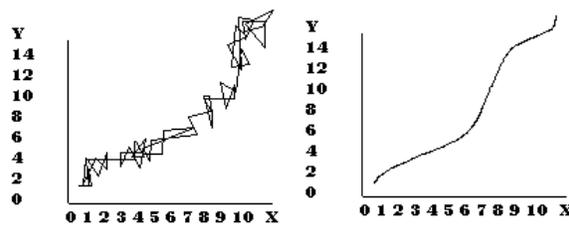

-Figure 6.8: Ink Drop Spread projection and extracted narrow path by Center of Gravity

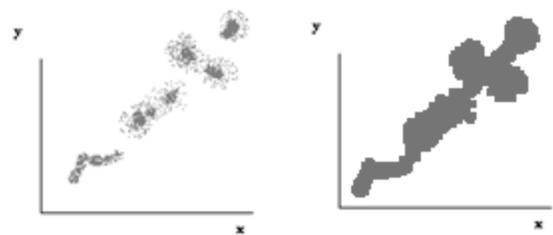

-Figure 6.9: same main data points and Extended Thickening

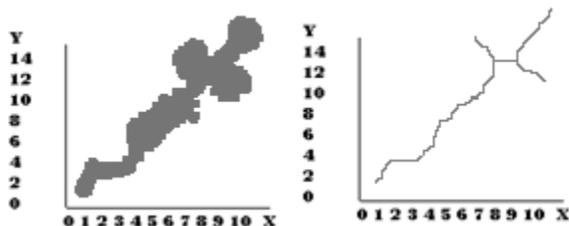

-Figure 6.10: Extended Thickening projection and extracted narrow path by Extended Thinning

The accuracy of narrow path and its details in comparison with IDS and COG are noticeable.

## 7. Conclusion

In this paper we reviewed ALM algorithm and introduced its operators, IDS and Center of Gravity. Then we explained that any Fuzzy modeling technique such as ALM should satisfy a few properties. Next it was shown that Center of Gravity does not satisfy the properties of a T-norm. To cope with this defect we introduced new operators which possessed three key aspects: capability of being applied in ALM, satisfying Fuzzy S-norm and T-norm conditions and being complements of each other. Operators were stated by mathematical expressions and the proof of their properties was presented.